\PassOptionsToPackage{colorlinks=true,allcolors=blue}{hyperref}

\documentclass[a4paper,fleqn]{cas-sc}

\usepackage[numbers]{natbib}

\usepackage{amsmath,amsfonts}

\usepackage{algorithm}
\usepackage{algpseudocode}

\usepackage{enumitem}

\usepackage{booktabs}
\usepackage{multirow}
\usepackage{array}

\usepackage[caption=false,font=normalsize,labelfont=sf,textfont=sf]{subfig}

\usepackage{textcomp}
\usepackage{stfloats}
\usepackage{url}
\usepackage{verbatim}
\usepackage{graphicx}
\usepackage{placeins}
\usepackage{xcolor}
\usepackage{wrapfig}
\usepackage{caption}
\usepackage{hyperref}
\hypersetup{colorlinks=true, citecolor=blue, linkcolor=blue, urlcolor=blue}
\usepackage{setspace}

\makeatletter
\@fleqnfalse
\makeatother

\newcommand{\AOP}{\varphi}      
\newcommand{\DOP}{\rho}         
\newcommand{\zen}{\theta}       
\newcommand{\az}{\phi}          

\algrenewcommand\algorithmicrequire{\textbf{Input:}}
\algrenewcommand\algorithmicensure{\textbf{Output:}}

\begin{document}

\let\WriteBookmarks\relax
\let\printorcid\relax

\shorttitle{Segmentation-Driven Monocular Shape from Polarization}
\shortauthors{J. Zhang et~al.}

\title[mode=title]{Segmentation-Driven Monocular Shape from Polarization
  based on Physical Model}


\author[1]{Jinyu~Zhang}

\author[1]{Xu~Ma}
\cormark[1]
\ead{maxu@bit.edu.cn}

\author[2]{Weili~Chen}

\affiliation[1]{%
  organization={Key Laboratory of Photoelectronic Imaging Technology and
    System of Ministry of Education of China,
    School of Optics and Photonics,
    Beijing Institute of Technology},
  city={Beijing},
  postcode={100081},
  country={China}}

\affiliation[2]{%
  organization={National Key Laboratory of Scattering and Radiation,
    Beijing Institute of Environmental Features},
  city={Beijing},
  postcode={100854},
  country={China}}

\cortext[cor1]{Corresponding author}

\begin{abstract}
Monocular shape-from-polarization (SfP) leverages the intrinsic
relationship between light polarization properties and surface geometry to
recover surface normals from single-view polarized images, providing a
compact and robust approach for three-dimensional (3D) reconstruction.
Despite its potential, existing monocular SfP methods suffer from azimuth
angle ambiguity---an inherent limitation of polarization analysis---that
severely compromises reconstruction accuracy and stability.
This paper introduces a novel segmentation-driven monocular SfP (SMSfP)
framework that reformulates global shape recovery into a set of local
reconstructions over adaptively segmented convex sub-regions.
Specifically, a polarization-aided adaptive region growing (PARG)
segmentation strategy is proposed to decompose the global convexity
assumption into locally convex regions, effectively suppressing azimuth
ambiguities and preserving surface continuity.
Furthermore, a multi-scale fusion convexity prior (MFCP) constraint is
developed to ensure local surface consistency and enhance the recovery of
fine textural and structural details.
Extensive experiments on both synthetic and real-world datasets validate
the proposed approach, showing significant improvements in disambiguation
accuracy and geometric fidelity compared with existing physics-based
monocular SfP techniques.
\end{abstract}



\maketitle
\doublespacing
\setlength{\abovedisplayskip}{10pt plus 2pt minus 5pt}
\setlength{\belowdisplayskip}{10pt plus 2pt minus 5pt}
\setlength{\abovedisplayshortskip}{0pt plus 3pt}
\setlength{\belowdisplayshortskip}{6pt plus 3pt minus 3pt}

\section{Introduction}

Three-dimensional (3D) reconstruction aims at recovering the stereoscopic
structures of objects from two-dimensional (2D) images~\cite{ref1}, with
wide applications in autonomous driving~\cite{ref2}, medical
diagnosis~\cite{ref3}, industrial manufacturing~\cite{ref4} and virtual
reality~\cite{ref5}.
Traditional methods such as stereo vision and structured light encounter
limitations of equipment complexity and lighting
sensitivity~\cite{ref6}.
Polarization-based 3D reconstruction, also named shape from polarization
(SfP), has emerged as a promising technique to solve the surface
geometries through polarization analysis.
SfP methods utilise the information of angle of polarization (AOP),
degree of polarization (DOP) and unpolarized intensity to recover the surface
normals from polarized images, offering the advantages of simplified
equipment, reduced lighting sensitivity and the capability of handling
transparent and reflective surfaces, making it a compelling computational
imaging modality for passive 3D sensing~\cite{ref7}.

However, traditional SfP methods encounter a critical challenge of
azimuth angle ambiguity~\cite{ref7}.
Due to the inherent aliasing problem of polarization analysis, surface
normal estimation often yields multiple possible solutions, thereby
significantly affecting reconstruction accuracy.
Existing solutions primarily include multi-modal fusion strategies (such
as combining depth sensors and photometric stereo), and deep learning
methods.
However, those approaches require either complex system structure or
extensive training data.
In addition, deep learning methods encounter the generalisation problem
in complex and ever-changing scenes.
In contrast, physics-based methods do not rely on training data and require minimal hardware, making them particularly suitable for
practical deployment~\cite{ref7}.
Thus, the physics-based SfP method is desired for practical
applications.

Monocular passive 3D reconstruction technology solves the surface normals
using only a single polarized image with unknown lighting condition,
evidently offering practical advantages.
However, real surfaces with diffuse reflection lack the one-to-one
correspondence between azimuth angle and AOP, leading to azimuth angle
ambiguity that is difficult to resolve without additional constraints.
Certain monocular SfP methods rely on a global convexity assumption to
address azimuthal angle ambiguity.
However, global convexity does not hold for objects with complex
structures, resulting in significant artefacts in the final
reconstruction.

To overcome this limitation, this paper proposes a fully physics-based
method, dubbed segmentation-driven monocular shape from polarization
(SMSfP), to resolve the azimuth angle ambiguity.
The key principle is reframing the global 3D reconstruction as a set of
independent reconstructions over locally convex sub-regions, thus
transforming the complex global problem into well-posed local ones.
In addition, a multi-scaled fusion convexity prior (MFCP) constraint is
proposed and applied in each sub-region to ensure the surface convexity
consistency, continuity, and texture clarity while avoiding abrupt
variations of surface normals, thereby suppressing azimuth angle
ambiguity and improving reconstruction accuracy.
The main contributions are summarised as follows:

\begin{enumerate}[topsep=0pt, partopsep=0pt, itemsep=2pt]
\item We propose a MFCP constraint, extracting textural details from the
  estimated azimuth angle to ensure local convexity and enhance
  reconstruction accuracy.
\item We propose a polarization-driven adaptive region growing (PARG)
  segmentation method that decomposes the global convexity assumption
  into a local convexity distribution~\cite{ref8}, ensuring surface
  continuity and thereby resolving azimuth angle ambiguity for complex
  object surfaces.
\item We propose the SMSfP framework employing the segmentation-driven
  reconstruction paradigm that integrates the above techniques.
  This approach demonstrates significant enhancement of disambiguation
  performance compared to other state-of-the-art physics-based monocular
  passive 3D reconstruction methods.
\end{enumerate}

\section{Related Works}

\subsection{Physics-based Methods}

Physics-based SfP methods can be categorised into two kinds of
approaches: the pure polarization-based methods and the multi-modal
fusion methods (SfP+X) that combine polarization states with additional
information sources.

\textbf{Pure polarization-based methods.}
Early research exploited polarization properties for 3D reconstruction
with significant limitations.
Drbohlav et al.\ reconstructed dielectric spheres but faced
inter-reflection constraints~\cite{ref9}.
Atkinson and Hancock applied diffuse polarization for shape
reconstruction, but found limited accuracy in regions away from object
boundaries~\cite{ref10}.
Miyazaki et al.\ addressed the azimuth ambiguity through target
rotation, requiring multiple image acquisitions from different
viewpoints~\cite{ref11}.
Additionally, Mahmoud et al.\ derived shading constraints from
polarization information for enhanced accuracy~\cite{ref12}.
Recent work by Smith et al.\ formulated SfP as an optimisation problem
of height estimation, achieving improved quality while remaining
vulnerable to azimuth ambiguities in complex
scenarios~\cite{ref13,ref14}.

\textbf{SfP + X.}
To overcome the limitations of pure polarization-based methods,
researchers combined polarization states with other complementary
information to alleviate the azimuth angle ambiguity.
Early work by Ngo Thanh et al.\ first integrated shading constraints for
small zenith angles~\cite{ref15}, while Atkinson and Hancock merged
polarization information with photometric stereo for enhanced
robustness~\cite{ref16,ref17}.
Stolz et al.\ used spectral imaging for transparent
objects~\cite{ref18}, and Morel et al.\ developed active illumination
systems for metallic surfaces~\cite{ref19}.

Recent approaches incorporated some modern sensing technologies.
Tozza et al.\ unified polarization and shading within the partial
differential equation frameworks~\cite{ref20}.
Kadambi et al.\ fused polarization with depth sensors~\cite{ref21}, and
Cui et al.\ developed polarimetric multi-view stereo~\cite{ref22}.
Additionally, Zhu et al.\ combined monocular SfP with a stereo cue from
an additional RGB camera~\cite{ref23}.
While these multi-modal approaches achieve superior reconstruction
performance, they require complex hardware setups that limit practical
deployment.

\subsection{Deep-learning-based Methods}

Deep learning has introduced powerful data-driven approaches for
polarization-to-geometry mapping.
Ba et al.\ pioneered deep SfP by integrating physical priors into neural
networks~\cite{ref24}, surpassing the traditional methods.
Recent work includes Lei et al.\ for outdoor scene
reconstruction~\cite{ref25}, Huang et al.\ for stereo polarization
systems~\cite{ref26}, and Lyu et al.\ for unknown illumination
scenarios~\cite{ref27}.
For specialised applications, Yang et al.\ designed underwater
de-scattering networks for turbid water reconstruction~\cite{ref28},
while Li et al.\ developed the SfP-U\textsuperscript{2}Net technique
significantly improving the accuracy of surface normal
estimation~\cite{ref29}.
More recently, learning-based methods have continued to advance across a range of scenarios, including unknown illumination~\cite{ref30},
attention-based architecture~\cite{ref31}, underwater scattering~\cite{ref32}, and sparse self-attention mechanism~\cite{ref33}, demonstrating strong performance on standard benchmarks and steadily improving reconstruction accuracy. Despite these advances, deep learning approaches generally require large-scale training data and substantial computational resources, and their data-driven nature leads to limited physical interpretability and poor generalisation to scenes outside the training distribution, limiting their practical applicability.

In contrast to existing methods that rely on complex hardware or large
datasets, this paper proposes a low-cost and fully physics-based
monocular framework that achieves competitive 3D reconstruction accuracy.

\section{Polarization Theory and Problem Formulation}

\subsection{Theoretical Foundation of Polarization}

Surface normal reconstruction requires establishing the equations that
relate normal vector components to measurable quantities.
Since the surface normal corresponds to the height gradients, the 3D
reconstruction problem reduces to height estimation.
The pixel component at coordinate $\mathbf{u}=(x,y)$ on a polarized image can be
calculated as follows~\cite{ref20}:
\begin{equation}
I_{\theta_j}(\mathbf{u})
=
\frac{I_{\max}+I_{\min}}{2}
+
\frac{I_{\max}-I_{\min}}{2}
\cos\!\big(2(\vartheta_j-\AOP(\mathbf{u}))\big),
\label{eq1}
\end{equation}
where $I_j(\mathbf{u})$ denotes the polarized intensity captured along the
angle of $\vartheta_j$; $I_{\max}$ and $I_{\min}$ denote the maximum
and minimum intensities measured over a full rotation of the polarizer;
$\AOP(\mathbf{u})$ represents the AOP of the scene.
The polarized image can be constructed from three parameters including
the AOP $\AOP$, the DOP $\DOP$, and the unpolarized intensity $I_{un}$~\cite{ref34},
where:
\begin{equation}
\DOP=\frac{I_{\max}-I_{\min}}{I_{\max}+I_{\min}},\qquad
I_{un}=\frac{I_{\max}+I_{\min}}{2}.
\label{eq2}
\end{equation}

Polarization state of light waves can be fully characterised by the
Stokes vector~\cite{ref22}.
The Stokes vectors can be expressed as a function of $\AOP$, $\DOP$ and 
$I_{un}$:
\begin{equation}
\hat{s} =
\begin{bmatrix}
s_0 \\ s_1 \\ s_2 \\ s_3
\end{bmatrix}
=
\begin{bmatrix}
I_0 + I_{90} \\
I_0 - I_{90} \\
I_{45} - I_{135} \\
0
\end{bmatrix}
=
\begin{bmatrix}
2I_{un} \\
2\rho \cos(2\varphi) \\
2\rho \sin(2\varphi) \\
0
\end{bmatrix},
\label{eq3}
\end{equation}
where $I_{0}$, $I_{45}$, $I_{90}$ and $I_{135}$ respectively represent
the intensities along the $0^\circ$, $45^\circ$, $90^\circ$ and
$135^\circ$ angles.
In addition, $I_{un}=s_0/2$ is the unpolarized intensity,
$\rho=\sqrt{s_1^2+s_2^2}/s_0$ is the DOP, and
$\varphi = \tan^{-1}(s_1/s_2)/2$ is the AOP.

\begin{figure}[htbp]\rmfamily
	\centering
	\includegraphics[width=\columnwidth]{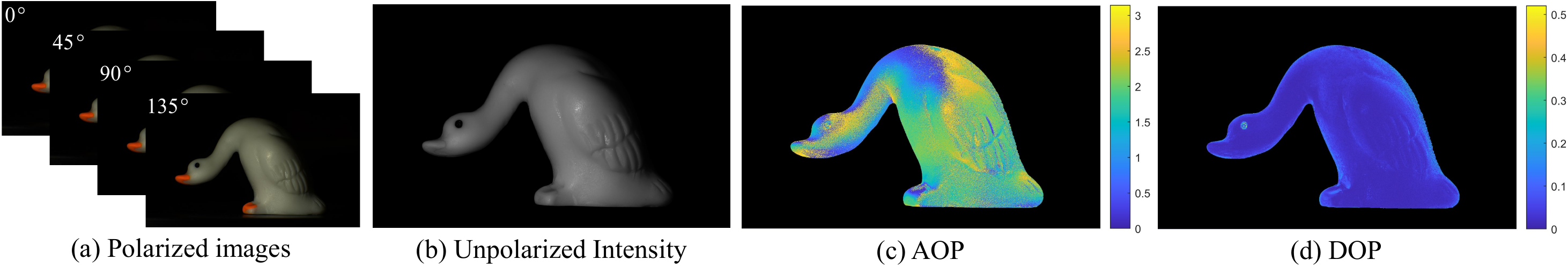}
	\caption{The polarized images and the decomposed components:
    (a)~polarized images along the angles of
    $0^\circ,45^\circ,90^\circ$ and $135^\circ$;
    (b)~unpolarized intensity image;} (c)~AOP image; (d)~DOP image.
	\label{fig1}
\end{figure}

Figure~\ref{fig1}(a) shows the polarized images of a swan figure along
the angles of $0^\circ$, $45^\circ$, $90^\circ$, $135^\circ$.
Figures~\ref{fig1}(b), \ref{fig1}(c) and \ref{fig1}(d) show the
corresponding unpolarized intensity $I_{un}$, AOP $\AOP$, and DOP $\DOP$
calculated from the four polarized images.

\subsection{Surface Normal Representation}

The normal surface vector $\hat{n}$ is parameterized by the zenith angle
and azimuth angle $\phi$ in spherical coordinates~\cite{ref14}:
\begin{equation}
\hat{n} =\big[\sin\zen\cos\az,\ \sin\zen\sin\az,\ \cos\zen\big]^{\!T},
\label{eq4}
\end{equation}
where $\zen$ is directly mapping to the DOP $\DOP$ and the refractive
index $\eta$, while $\phi$ relates to the AOP $\AOP$.
However, azimuth ambiguity and measurement noise bring difficulties to
the direct estimation of $\phi$.
In order to overcome this problem, an alternative method can be used.

Let $z$ denote the unknown surface height.
Rather than computing the surface normal $\hat{n}$ directly from the
azimuth angle $\phi$, we can establish constraints on the gradient of
height, $\nabla z$, which is directly related to $\hat{n}$ as
follows~\cite{ref20}:
\begin{equation}
\hat{n} = \frac{[-z_x, -z_y, 1]^{\top}}{\sqrt{z_x^2 + z_y^2 + 1}}
= \frac{[-\nabla z, 1]^{\top}}{\sqrt{1 + | \nabla z |^2}},\quad
\nabla z = [z_x, z_y],
\label{eq5}
\end{equation}
where $z_x$ and $z_y$ denote the partial derivatives of $z$ with respect
to the $x$ and $y$ directions, respectively.

\subsection{Diffuse Polarization Model}

To estimate zenith angle, we adopt the diffuse polarization model,
assuming all pixels are dominated by diffuse reflection.
This model assumes that the polarization is caused by the light
scattering from the subsurface and the subsequent Fresnel transmission
upon exiting the surface~\cite{ref10}.
For diffuse reflection, the zenith angle relates directly to the
DOP~\cite{ref14}:
\begin{equation}
\rho = \frac{
    \left( \eta - \frac{1}{\eta} \right)^2 \sin^2\theta
}{
    2 + 2\eta^2 - \left( \eta + \frac{1}{\eta} \right)^2 \sin^2\theta
    + 4\cos\theta\, \sqrt{\eta^2 - \sin^2\theta}
},
\label{eq6}
\end{equation}
where $\eta$ represents the refractive index.
Therefore, we can derive the equation for $\theta$ as follows:
\begin{equation}
\label{eq7}
\begin{split}
    \cos(\theta) = \sqrt{\frac{\eta^4(1-\rho^2) + 2\eta^2(2\rho^2+\rho-1) + \rho^2 + 2\rho - 4\eta^3\rho\sqrt{1-\rho^2} + 1}{(\rho+1)^2(\eta+1) + 2\eta^2(3\rho^2+2\rho-1)}}.
\end{split}
\end{equation}

\section{Proposed Method}

To address the problem of azimuth angle ambiguity, we propose the SMSfP
method as shown in Fig.~\ref{fig2}.
The workflow proceeds as follows:

\begin{figure}[htbp]\rmfamily
  \centering
  \includegraphics[width=\textwidth]{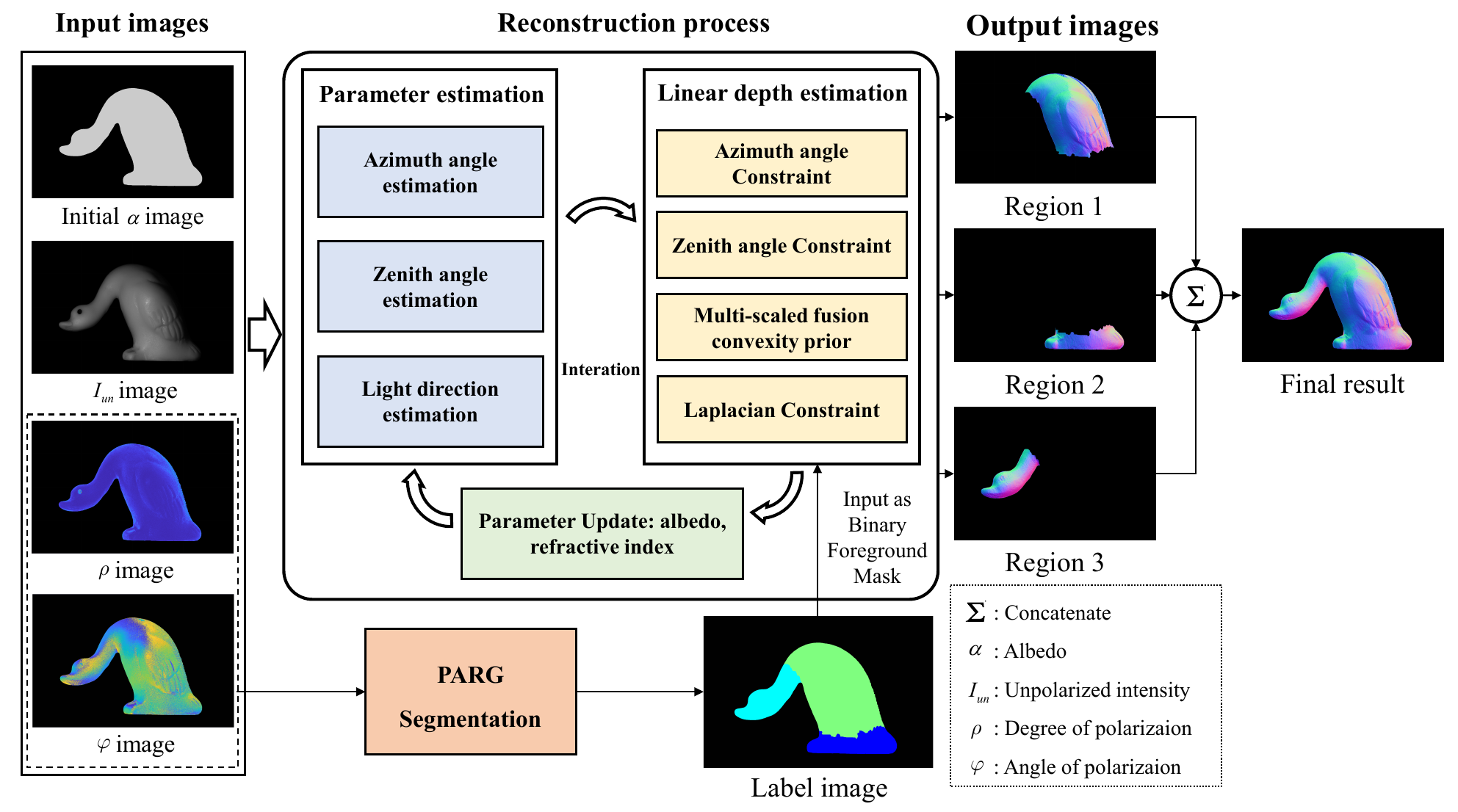}
  \caption{The diagram of the proposed SMSfP method.}
  \label{fig2}
\end{figure}

\begin{enumerate}
\item \textbf{Input data.}
  Input the initial albedo $\alpha$, unpolarized intensity $I_{un}$, DOP $\rho$
  and AOP $\varphi$.

\item \textbf{Segment each sub-region.}
  Use the PARG segmentation method to obtain the binary foreground mask
  for each sub-region.

\item \textbf{Shape reconstruction with constraints.}
  Reconstruct each sub-region independently via iterative optimisation
  using the zenith angle, azimuth angle, MFCP, and Laplacian
  constraints.

\item \textbf{Post-processing.}
  Concatenate the reconstruction results of sub-regions and use guided
  filter to smooth the stitching boundaries~\cite{ref35}.
\end{enumerate}

\subsection{Azimuth Angle Constraint}

For a diffuse reflection-dominated pixel, its azimuth angle exhibits
inherent ambiguity with two possible values differing by
$\pi$~\cite{ref14}.
The projection of $\hat{n}$ onto the $x$-$y$ plane is parallel to the
azimuth direction, allowing both possible azimuth angles to satisfy the
geometric constraints.
This condition is expressed as (assuming azimuth angle
$\phi = \varphi \pm \pi$)~\cite{ref13}:
\begin{equation}
\hat{n}\cdot[\cos\az,\ -\sin\az,\ 0]^{T}=0.
\label{eq8}
\end{equation}

Using the height gradient $\nabla z$ in Eq.~(\ref{eq5}),
Eq.~(\ref{eq8}) can be rewritten as a height constraint on $z$ in terms
of $\phi$:
\begin{equation}
[-\cos\phi,\ \sin\phi,\ 0]^T\cdot\nabla z = 0.
\label{eq9}
\end{equation}

\subsection{Zenith Angle Constraint}

The zenith angle constraint relates $\theta$ to $z$ through the viewing
direction $\hat{v}$.
The relationship between $\theta$ and normal $\hat{n}$ is~\cite{ref20}:
\begin{equation}
\cos\zen=\hat{n}\!\cdot\!\hat{v}=\frac{-\nabla z \cdot [v_1,\ v_2]^T+v_3}{\sqrt{1+|\nabla z|^2}},
\label{eq10}
\end{equation}
where $\hat{v} = [v_1,v_2,v_3]^T$ represents the viewing direction.

The unpolarized intensity $I_{un}$ offers a further constraint on the surface
orientation based on Lambert's law, a reflectance model that describes
the ideal diffuse reflection, where the light is scattered uniformly in
all directions~\cite{ref20}.
The relationship between $I$ and $\hat{n}$ is given by ~\cite{ref20}:
\begin{equation}
I_{un}=\alpha\,\hat{n}\!\cdot\!\hat{l}=\alpha\,\frac{-\nabla z \cdot [l_1,\ l_2]^T + l_3}{\sqrt{1+|\nabla z|^2}},
\label{eq11}
\end{equation}
where $\alpha$ and $\hat{l}=[l_1,l_2,l_3]^T$ represent the albedo and
illumination direction, respectively.
Using the common term $\sqrt{1+|\nabla z|^2}$ as an intermediate
equality between Eqs.~(\ref{eq10}) and (\ref{eq11}), we can derive:
\begin{equation}
\frac{-\nabla z \cdot [v_1,v_2]^T+v_3}{\cos \theta}=\alpha\frac{-\nabla z \cdot [l_1,\ l_2]^T + l_3}{I_{un}},
\label{eq12}
\end{equation}
where $\hat{l} \neq \hat{v}$, $\cos \theta$ and $I_{un} \neq 0$.
Illumination direction $\hat{l}$ is estimated by the method proposed
in~\cite{ref14}.

\subsection{Multi-Scaled Fusion Convexity Prior}

In addition to the azimuth and zenith angles, the object's mask also
offers additional geometric constraints for surface reconstruction.
To exploit this information, Smith et al.\ proposed a convexity prior
constraint that derives additional azimuth angles from mask
boundaries~\cite{ref14}.
We refer to those additional azimuth angles as implicit azimuth angles
$\phi_{im}$ throughout this paper.
Specifically, the $\phi_{im}$ are computed by assuming the global object
convexity and using geometric propagation methods.
The computation employs the mask erosion or closest-boundary assignment
to propagate the boundary orientation information inward throughout the
object interior~\cite{ref36,ref37}.
The resulting azimuth-angle estimates are then combined with the zenith
angles to construct the outward-pointing prior normals.
However, this method has several limitations.
It yields the implicit angles with limited accuracy and exhibits
spatially discretised distribution.
Moreover, it fails to capture the surface texture variations since it
relies solely on the mask shape.

\begin{figure}[htbp]\rmfamily
  \centering
  \includegraphics[width=\textwidth]{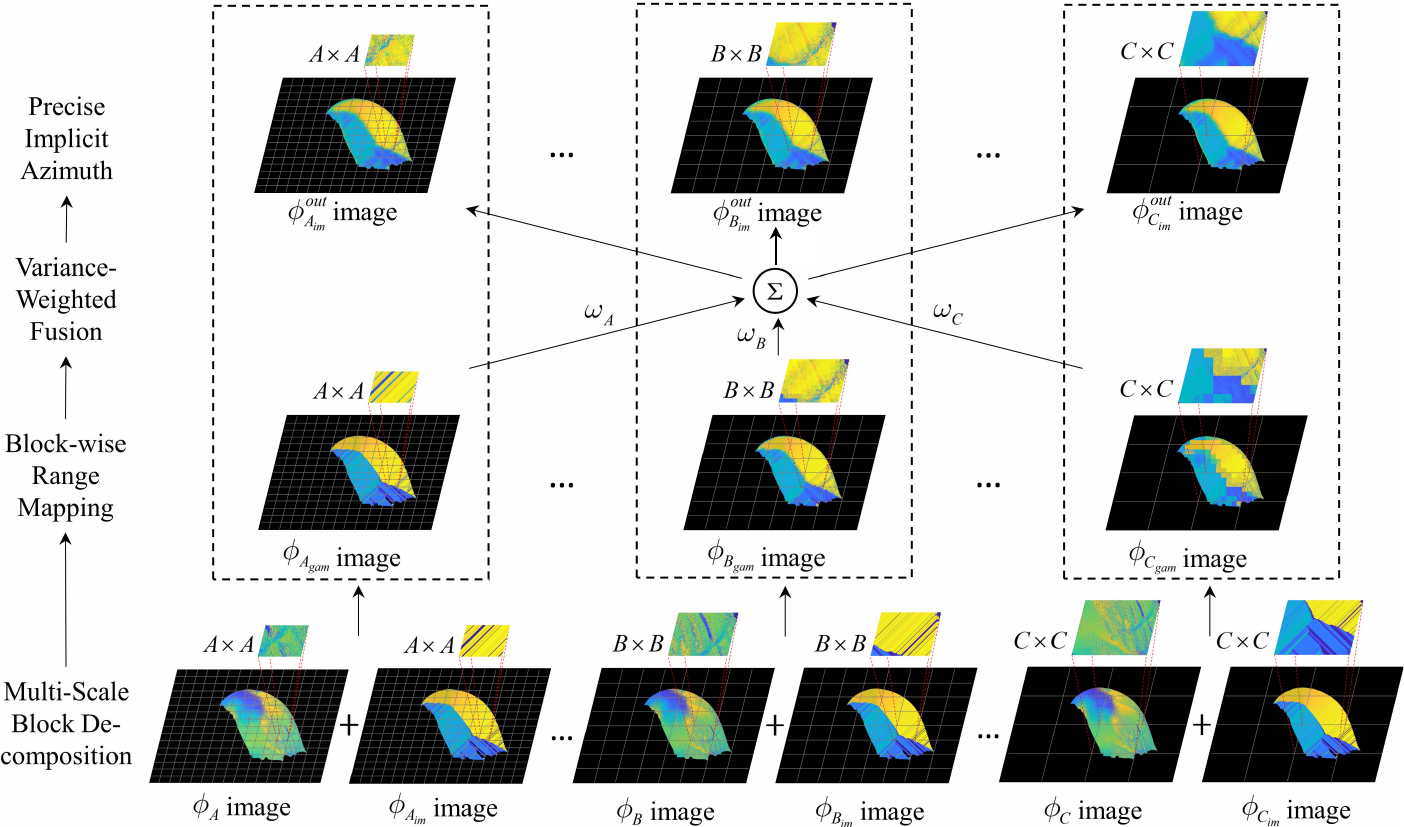}
  \caption{Computational workflow for multi-scale fusion convexity prior
    constraint.}
  \label{fig3}
\end{figure}

To overcome those limitations, we propose a multi-scaled fusion
framework for improving the implicit azimuth angle estimation with
richer textural details (as shown in Fig.~\ref{fig3}).
The proposed framework extracts multi-scaled features from the estimated
azimuth angles through variance-weighted fusion.
This enables the incorporation of textural details from estimated azimuth
angles, while maintaining the prior distribution properties of the
implicit azimuth angles.

The workflow in Fig.~\ref{fig3} proceeds as follows.

\textbf{Multi-Scaled Block Decomposition.}
We decompose both azimuth angles (assume that $\phi = \varphi$) and
implicit azimuth angles into blocks with different sizes of
$A \times A$, \ldots, $B \times B$, \ldots, $C \times C$.
This yields block-decomposed angles at each scale:
$[\phi_A,\ \phi_{A_{im}}]$, \ldots, $[\phi_B,\ \phi_{B_{im}}]$,
\ldots, $[\phi_C,\ \phi_{C_{im}}]$.

\textbf{Block-wise Range Mapping.}
We first linearly normalise each block in the block-decomposed azimuth
angles $(\phi_A,\ldots,\phi_B,\ldots,\phi_C)$ to $[0,\ 1]$, then apply
gamma transformation ($\gamma = 0.5$) in each block.
The gamma-transformed azimuth angles
$\phi_{A_{gam}} = (\phi_A)^\gamma$, \ldots,
$\phi_{B_{gam}} = (\phi_B)^\gamma$, \ldots,
$\phi_{C_{gam}} = (\phi_C)^\gamma$ are then mapped to match the value
ranges of the corresponding implicit azimuth angles
$\phi_{A_{im}}$, \ldots, $\phi_{B_{im}}$, \ldots, $\phi_{C_{im}}$.
This process preserves the value distribution of implicit azimuth
angles, while incorporating the detailed features from azimuth angles.

\textbf{Variance-Weighted Fusion.}
We calculate the variance of azimuth angles at each scale, yielding
$\sigma_A$, \ldots, $\sigma_B$, \ldots, $\sigma_C$ respectively.
Those  variances  serve  as  weight  coefficients  inthe summation process to compute the final implicit azimuthangles. The calculations are given by:
\begin{equation}
\phi_{im}^{out}=\sum_{i\in\{A,\ldots,C\}}\omega_i\,\phi_{i_{im}},\qquad
\omega_i = \frac{\sigma_i^2}{\sum_{j} \sigma_j^2},
\label{eq13}
\end{equation}
where $\omega_i$ are the weight coefficients for the implicit azimuth
angles at scale $i$.
After obtaining the fused implicit azimuth angles $\phi^{out}_{im}$, we
can use them and the zenith angles to construct the implicit normal
vectors $\hat{n}_{im}$ as priors:
\begin{equation}
\hat{n}_{im}=\big[\sin\zen\cos\phi_{\mathrm{im}}^{\mathrm{out}},\
\sin\zen\sin\phi_{\mathrm{im}}^{\mathrm{out}},\ \cos\zen\big]^{\!T}.
\label{eq14}
\end{equation}

Combining Eq.~(\ref{eq4}) and Eq.~(\ref{eq5}), we can derive the
partial derivatives of height along the $x$ and $y$ directions as
follows:
\begin{equation}
z_x=\frac{-\sin\theta \cos \phi}{\cos\theta},\qquad
z_y=\frac{-\sin\theta \sin \phi}{\cos\theta}.
\label{eq15}
\end{equation}

We first substitute Eq.~(\ref{eq15}) into Eq.~(\ref{eq4}).
Then, the $z_x$ and $z_y$ are replaced by the finite difference
gradient operators $\mathbf{D} = [\mathbf{D}_x, \mathbf{D}_y]^T$
applied to $z$, yielding the estimated normal vector $\hat{n}_{est}$:
\begin{equation}
\hat{n}_{est}=[-\mathbf{D}_x*z\cos\theta,\ -\mathbf{D}_y*z\cos\theta,\ \cos\theta]^T,
\label{eq16}
\end{equation}
where $z_x = \mathbf{D}_x*z$ and $z_y = \mathbf{D}_y*z$.
We adopt the weighting scheme from Smith et al., applying adaptive
weights $\omega_{con}(\mathbf{u})$ to the convexity prior~\cite{ref14}.
These weights range from 0 to 1, with maximum values at boundary pixels
and decaying exponentially toward the interior regions.

Therefore, the optimisation loss function for the MFCP constraint takes
the form:
\begin{equation}
\mathcal{E}_{convex}(z)=\sum_y\sum_{x}\omega_{con}^2(\mathbf{u})\big\|\hat{n}_{im}(\mathbf{u})-\hat{n}_{est}(\mathbf{u})\big\|^2,
\label{eq17}
\end{equation}
where the weighted terms enforce the alignment between the estimated
normals $\hat{n}_{est}$ and the implicit normals $\hat{n}_{im}$.

\subsection{Height Estimation and Iterative Parameter Update}

Based on the constraints in Sections~4.1 to 4.3 and the Laplacian
constraint that enforces surface smoothness by minimising height
variations between neighbouring pixels~\cite{ref14}, we formulate a
linear least-squares problem to solve for the final height map.
Following the formulation in~\cite{ref14}, the problem is cast as
minimising the objective function:
\begin{equation}
\mathcal{E}(z)=\big\|\mathbf{A} \mathbf{D} z - b\big\|^2,
\label{eq18}
\end{equation}
where the matrix $\mathbf{A}$ incorporates the coefficients of height
gradient, $\mathbf{D}z$, from the constraints, and the vector
$\mathbf{b}$ represents the constant terms of those constraints.
To discretise the height derivatives, we employ a Gaussian-smoothed
central difference scheme, which adapts at boundaries by reverting to
the simpler finite differences.
The resulting large and sparse linear system is then solved using the QR
decomposition.

After obtaining the initial estimation of $z$, we employ the least
squares again to update the albedo and refractive index $\eta$:
\begin{equation}
\min_{\alpha,\eta}\ \sum_{x}\big\|\rho_{est}-\rho_d(\theta,\eta)\big\|^2,
\label{eq19}
\end{equation}
where $\rho_{est}$ denotes the estimated DOP calculated from
Eq.~(\ref{eq3}), and $\rho_d$ is the DOP computed from the estimated
surface via Eq.~(\ref{eq6}).
After updating $\alpha$ and $\eta$, we update the zenith angle $\theta$
using Eq.~(\ref{eq7}).
Then, Eq.~(\ref{eq18}) is solved iteratively until $z$ converges.

\subsection{Polarization-Driven Region Segmentation}

As established in Section~4.3, the MFCP constraint is used for
globally convex objects.
However, for complex objects with multiple local convex regions,
applying this constraint to the entire foreground mask will introduce
significant reconstruction errors.
To solve this problem, we propose a polarization-driven adaptive region
growing (PARG) segmentation method to partition the entire object
surface into a set of locally convex segments~\cite{ref8}.
Each of those sub-regions can then be processed independently,
effectively decomposing the challenging global reconstruction problem
into a set of manageable local problems.

\begin{figure}[htbp]\rmfamily
  \centering
  \includegraphics[width=0.9\columnwidth]{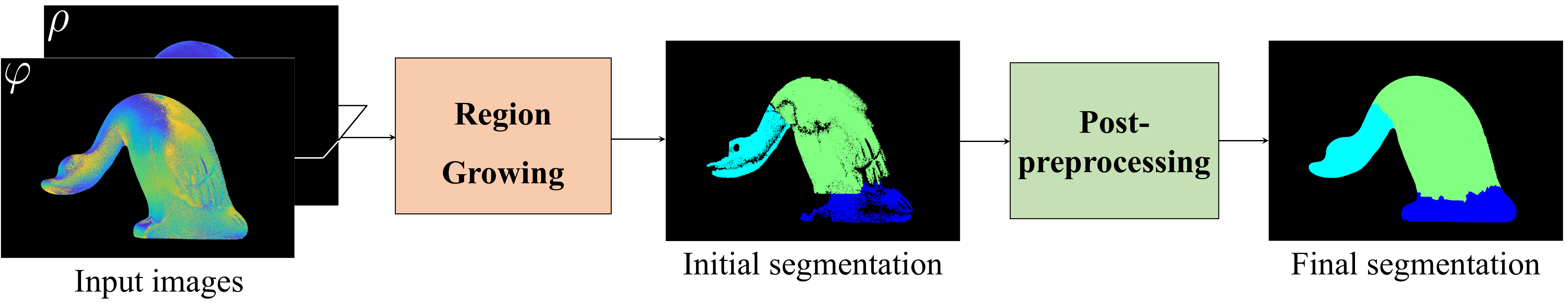}
  \caption{The workflow of PARG segmentation method, where the region
    growing is applied to the input DOP $\rho$ and AOP $\varphi$, and
    the post-processing is used to generate the final segmentation.}
  \label{fig4}
\end{figure}

Algorithm~\ref{alg1} outlines the complete procedure, where a
pre-defined similarity threshold, $\tau$, governs the growing criterion,
and Fig.~\ref{fig4} shows the workflow.
The algorithm operates on a four-dimensional (4D) feature tensor derived
from the polarization cues, constructed as follows:
\begin{equation}
\mathbf{F}(\mathbf{u}) =
\begin{bmatrix}
\rho(\mathbf{u}) \\
\cos 2\varphi(\mathbf{u}) \\
\sin 2\varphi(\mathbf{u}) \\
|\nabla \varphi(\mathbf{u})|
\end{bmatrix},\qquad
|\nabla \varphi| = \sqrt{
    \left( \frac{\partial \varphi}{\partial x} \right)^2
    + \left( \frac{\partial \varphi}{\partial y} \right)^2
},
\label{eq20}
\end{equation}
where $\rho$, $\varphi$ and $|\nabla \varphi|$ represent the DOP, AOP
and the gradient magnitude of AOP, respectively.
The employment of $\cos 2\varphi(\mathbf{u})$ and $\sin 2\varphi(\mathbf{u})$
eliminates the periodicity issue in the polarization angles by ensuring
that equivalent angles (differing by $\pi$) produce identical feature
values, thus mitigating trigonometric periodicity interference in the
segmentation.
The $\rho$ and $|\nabla \varphi|$ are utilised because their variations
correlate with the surface geometry, providing effective boundary
information for neighbourhood scanning.

The region growing method produces the initial segmentation result.
Then, we refine the segmentation by post-processing techniques including
the morphological reconstruction-based hole filling and Gaussian
filtering for boundary smoothing~\cite{ref38}.
Each labelled region corresponds to a locally convex sub-region and
provides a binary mask for the independent reconstruction as described
in Section~4.4.

As shown in Algorithm~\ref{alg1}, our approach follows the standard
region growing framework, which typically involves weight calculation
and feature distance computation at each iteration.
Our key contributions lie in enhancing these two core components, that
is, introducing adaptive weight calculation based on the local variance,
and developing a 4D feature distance computation based on polarization
cues, as detailed below.

\begin{algorithm}[t]
\caption{Polarization-Driven Region Growing Segmentation}
\label{alg1}
\small
\begin{algorithmic}[1]
\Require $\rho, \varphi, M(\text{mask}), \lambda_{\rho}, \lambda_{\varphi}, \tau$
\Ensure $L(\text{labels})$
\Statex
\State \textit{// 1. Initialization}
\State $\mathbf{F} \gets [\rho, \sin(2\varphi), \cos(2\varphi), |\nabla\varphi|]^T$
\State $S \gets \text{InitializeSeeds}(M)$, $L \gets \emptyset$, $Q \gets S$
\Statex
\State \textit{// 2. Adaptive Region Growing}
\While{$Q \neq \emptyset$}
    \State $p \gets \text{dequeue}(Q)$
    \For{each neighbor $q$ of $p$}
        \If{$L[q] = 0$ \textbf{and} $M[q] = 1$}
            \State \textit{// Adaptive Weight Calculation}
            \State $\sigma_{\rho} \gets \sigma(\rho, W_{5\times 5}(q))$
            \State $\sigma_{\varphi} \gets \sigma(\varphi, W_{5\times 5}(q))$
            \State $R_{\rho} \gets \exp(-\sigma_{\rho}^{2}/\max(\sigma_{\rho}^{2}))$
            \State $R_{\varphi} \gets \exp(-\sigma_{\varphi}^{2}/\max(\sigma_{\varphi}^{2}))$
            \State $\mathbf{W} \gets [1+\lambda_{\rho}R_{\rho},\, 1+\lambda_{\varphi}R_{\varphi},\, 1+\lambda_{\varphi}R_{\varphi},\, 1]^T$
            \Statex
            \State \textit{// Weight Distance Calculation}
            \State $\mathbf{F}_{seed} \gets \text{GetSeedFeatureForRegion}(L(p))$
            \State $d_{feature} \gets \lVert \mathbf{W} \odot (\mathbf{F}_{neighbor} - \mathbf{F}_{seed}) \rVert_2$
            \If{$d_{feature} < \tau$}
                \State $L(q) \gets L(p)$, $\text{enqueue}(Q, q)$
                \State $\text{UpdateSeedFeature}(L(p), \mathbf{F}_q)$
            \EndIf
        \EndIf
    \EndFor
\EndWhile
\Statex
\State \textit{// 3. Post-processing}
\State $L \gets \text{PostProcess}(L)$
\State \Return $L$
\end{algorithmic}
\end{algorithm}

\textbf{Adaptive Weight Calculation.}
Weight calculation generates a vector of weight coefficients based on
the local stability of polarization features around a candidate pixel.
The weight coefficients adaptively modulate the importance of each
feature channel in the subsequent distance calculation.
For each neighbouring pixel under examination, the algorithm first
computes the variances of DOP and AOP within the $5\times5$ local
window:
\begin{equation}
\sigma^2_{\rho}(\mathbf{u})=\sigma\big(\rho_{W_{5 \times 5}(\mathbf{u})}\big),
\label{eq21}
\end{equation}
\begin{equation}
\sigma^2_{\varphi}(\mathbf{u})=\sigma\big(\varphi_{W_{5 \times 5}(\mathbf{u})}\big),
\label{eq22}
\end{equation}
where $\sigma$ and $W_{5\times 5}(\mathbf{u})$ represent the variance
calculation function and the $5\times5$ window centred at pixel
$\mathbf{u}$.

The adaptive weight calculation employs a variance-based reliability
assessment to dynamically adjust feature importance.
For each pixel $\mathbf{u}$, we first compute the reliability scores based
on local variances:
\begin{equation}
R_\rho(\mathbf{u})=\exp\!\Big(-\sigma^2_{\rho}(\mathbf{u})/\max\big(\sigma^2_{\rho}\big)\Big),
\label{eq23}
\end{equation}
\begin{equation}
R_\varphi(\mathbf{u})=\exp\!\Big(-\sigma^2_{\varphi}(\mathbf{u})/\max\big(\sigma^2_{\varphi}\big)\Big),
\label{eq24}
\end{equation}
where $R_{\rho}$ and $R_{\varphi}$ represent the reliability scores of
DOP and AOP, respectively, with higher values indicating more reliable
features that will receive larger weights.
The adaptive weight vector is then computed as:
\begin{equation}
\mathbf{W}(\mathbf{u}) =
\begin{bmatrix}
w_\rho \\
w_{\cos} \\
w_{\sin} \\
w_g
\end{bmatrix}
=
\begin{bmatrix}
1 + \lambda_\rho R_\rho \\
1 + \lambda_\varphi R_\varphi \\
1 + \lambda_\varphi R_\varphi \\
1
\end{bmatrix},
\label{eq25}
\end{equation}
where $\lambda_{\rho}$ and $\lambda_{\varphi}$ control the adaptive
strengths.

\textbf{4D Feature Distance.}
The 4D weighted distance calculation uses the adaptive weight vector
$\mathbf{W}(\mathbf{u})$ to compute a final dissimilarity score, which serves
as the decision metric for merging the pixels into different regions.
Feature distance computation employs the adaptive weighted Euclidean
distance~\cite{ref39}, where the feature difference vector is first
element-wisely multiplied by the adaptive weight vector, followed by the
$\mathcal{L}_2$ norm calculation:
\begin{equation}
d_{feature}=\big\|\mathbf{W}(x,y)\odot\big(\mathbf{F}_{neighbor}-\mathbf{F}_{seed}\big)\big\|_2,
\label{eq26}
\end{equation}
where $F_{neighbor}$ and $F_{seed}$ represent the feature vectors of
the neighbouring pixel and the seed pixel, respectively; $\odot$ denotes
the element-wise multiplication~\cite{ref40}.
These weights emphasise more reliable features while de-emphasising less
reliable ones.

\section{Experiment and Analysis}

This section presents a comprehensive experimental validation of the
proposed method.
Section~5.1 introduces the two synthetic datasets together with the
unified parameter settings and evaluation metrics adopted throughout
the experiments.
In Section~5.2, we conduct a quantitative comparison on two synthetic
datasets against three monocular passive reconstruction algorithms:
Atkinson et al.~\cite{ref10}, Mahmoud et al.~\cite{ref12}, and Smith et
al.~\cite{ref14}.
Subsequently, Section~5.3 provides an ablation study to validate the
impact of PARG segmentation method.
Finally, Section~5.4 introduces the real-world dataset and the
polarized imaging testbed built by our group, and compares the
algorithm's performance against existing methods using the real-world
data.

\subsection{Synthetic Datasets and Experimental Settings}

The following simulation experiments use three datasets: a synthetic data (noted
as dataset A), the Deschaintre's dataset (noted as dataset B)~\cite{ref41}.
Dataset A contains four objects (camera, bird, car, teapot) created from
publicly available 3D models of Sketchfab and rendered using Adobe
Substance 3D Painter with the material model from Deschaintre et al.\ at
$1024\times1024$ resolution~\cite{ref42}.
Dataset B contains four objects (dog, human, sheep, cup) synthesised
using the same methodology at $512\times512$ resolution.

Across all experiments, we set the initial albedo $\alpha = 0.8$, view
direction $\hat{v} = [0,0,1]^T$, initial refractive index $\eta = 1.15$,
and the PARG's adaptive weights $\lambda_{\rho}$ and $\lambda_{\varphi}
= 2$.
The albedo and refractive index are empirical values suited for common diffuse materials, and are adaptively refined through the iteration in Eq. (19) to accommodate material variations. The view direction assumes orthographic projection. The PARG adaptive weights are determined empirically and found to be robust across the tested datasets.
Furthermore, the reconstruction performance is assessed using the mean
angular error (MAE) and root mean square error (RMSE) of angles between
the estimated normals and the ground truth (GT)~\cite{ref26}.
We also calculate the percentage of pixels with angular errors under the
thresholds of $11.25^\circ$, $22.5^\circ$ and $30.0^\circ$~\cite{ref43},
denoted as the ($11.25^\circ$/$22.5^\circ$/$30.0^\circ$) pixel accuracy.

\subsection{Experimental Results on Synthetic Data}

\begin{figure}[!b]\rmfamily
  \centering
  \includegraphics[width=\textwidth]{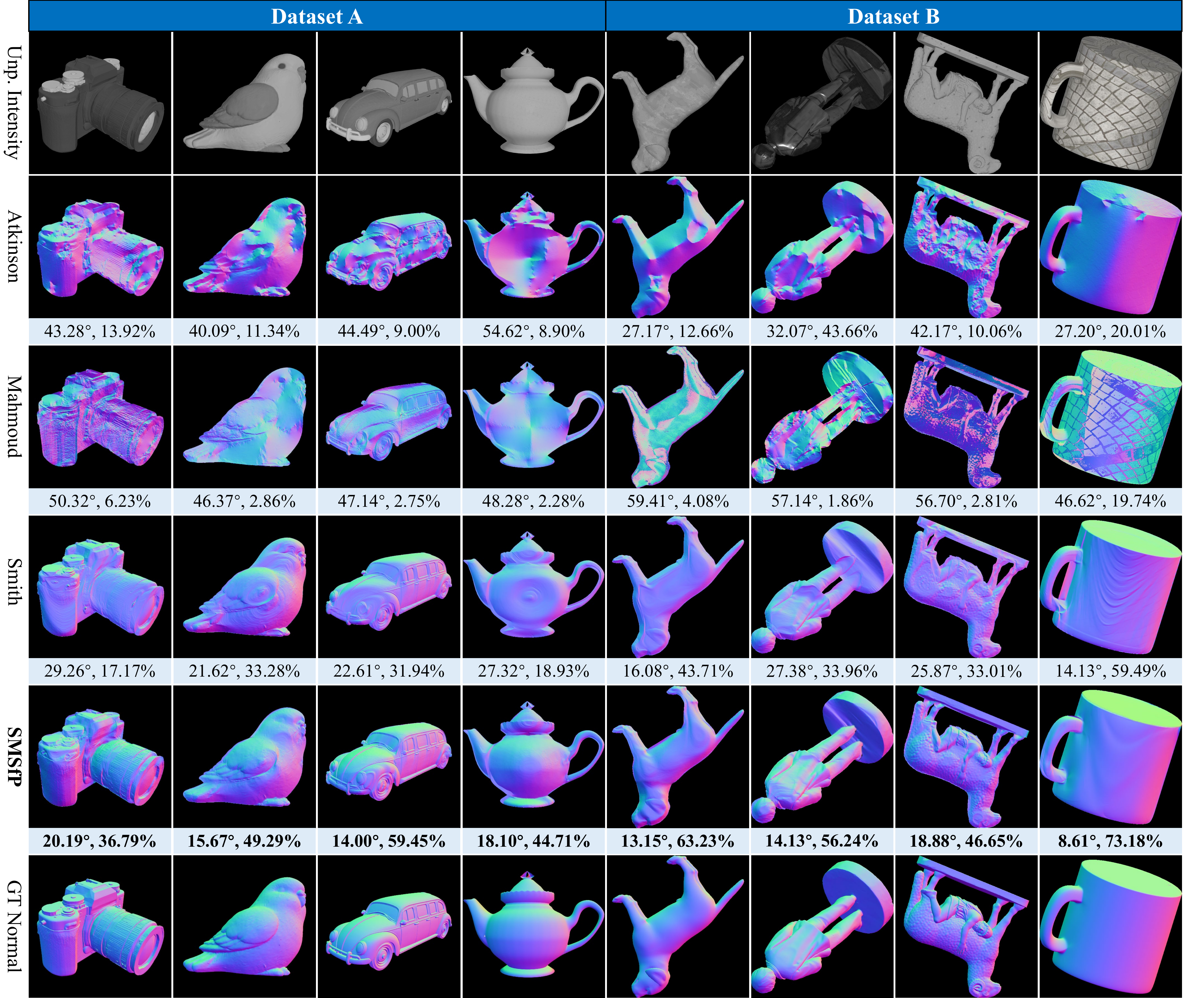}
  \caption{Performance comparison of the proposed SMSfP method against
    the baseline methods on dataset A and dataset B.
    From top to bottom, the rows display: the input unpolarized intensity
    images, results obtained by the baseline methods
    (Atkinson~\cite{ref10}, Mahmoud~\cite{ref12} and
    Smith~\cite{ref14}), the proposed SMSfP method, and the GT normal
    maps.
    The numbers below each result indicate the MAE in degrees and the
    pixel accuracy ($<11.25^\circ$), respectively.}
  \label{fig5}
\end{figure}

Figure~\ref{fig5} presents the reconstruction results on dataset A and
dataset B.
The comparative analysis demonstrates the superior performance of our
proposed method across diverse object geometries.
The first row shows the input unpolarized intensity images.
The second to the fifth rows display the reconstruction results obtained
by the Atkinson's method~\cite{ref10}, Mahmoud's
method~\cite{ref12}, Smith's method~\cite{ref14}, and the proposed
method, respectively.
The GT normal maps are shown in the bottom row.

\textbf{The results of dataset A.}
The baseline methods show significant limitations, where the Atkinson's
and Mahmoud's methods achieve MAE of $40.09^\circ$--$54.62^\circ$ and
exhibit noisy and discontinuous reconstructed maps.
The Smith's method improves the performance (MAE: $21.62^\circ$--$29.26^\circ$),
but retains artefacts in the regions with complex geometries.
Our approach demonstrates superior reconstruction quality with MAE
reduced to $14.00^\circ$--$20.19^\circ$ and substantially improved
pixel accuracy (36.79\%--59.45\% at $11.25^\circ$ threshold), while
maintaining smooth surface continuity and fine structural details.

\textbf{The results of dataset B.}
The baseline methods show some variability, with the Atkinson's and
Mahmoud's methods achieving MAE of $27.17^\circ$--$42.17^\circ$ and MAE
of $46.62^\circ$--$59.41^\circ$, respectively.
The Smith's approach provides better accuracy
(MAE: $14.13^\circ$--$27.38^\circ$) but shows limitations when dealing
with complex objects.
Our method consistently achieves the best performance across all objects
(MAE: $8.61^\circ$--$18.88^\circ$) with substantially higher pixel
accuracy (46.65\%--73.18\%), effectively reconstructing the challenging
geometries and complex structural arrangements.

Table~\ref{tab1} presents comprehensive quantitative results, where
``*'' indicates the best performance under a certain evaluation metric
and the performance of our method is indicated in bold.
Our method substantially outperforms all baselines, respectively
achieving MAEs of $16.99^\circ$ and $13.69^\circ$ on datasets A and B,
representing $8.21^\circ$ and $7.18^\circ$ improvements over the best baseline
method (the Smith's method).
Consistent advantages are observed across all thresholds, with pixel
accuracy reaching 47.56\%--59.83\% at $11.25^\circ$ and 80--90\% at
higher thresholds.
These improvements stem from the synergistic combination of PARG
segmentation method and MFCP constraint, which effectively mitigate
azimuth ambiguities inherent in the traditional approaches.

\begin{table}[htbp]\rmfamily
  \centering
  \caption{Quantitative comparison of all methods on dataset A and
    dataset B~\cite{ref41}.}
  \label{tab1}
  \footnotesize
  \setlength{\tabcolsep}{4pt}
  \begin{tabular}{@{} l ccccc ccccc @{}}
    \toprule
    \multirow{3}{*}{Method}
      & \multicolumn{5}{c}{Dataset A}
      & \multicolumn{5}{c}{Dataset B} \\
    \cmidrule(lr){2-6}\cmidrule(lr){7-11}
    & \multicolumn{2}{c}{Angular Error (deg.)}
      & \multicolumn{3}{c}{Pixel Accuracy (\%)}
      & \multicolumn{2}{c}{Angular Error (deg.)}
      & \multicolumn{3}{c}{Pixel Accuracy (\%)} \\
    \cmidrule(lr){2-3}\cmidrule(lr){4-6}\cmidrule(lr){7-8}\cmidrule(lr){9-11}
    & MAE & RMSE & $11.25^\circ$ & $22.5^\circ$ & $30^\circ$
    & MAE & RMSE & $11.25^\circ$ & $22.5^\circ$ & $30^\circ$ \\
    \midrule
    Atkinson & 45.86 & 50.27 & 10.79 & 32.73 & 43.43
             & 32.15 & 36.77 & 21.60 & 41.47 & 57.93 \\
    Mahmoud  & 48.03 & 50.44 &  3.53 & 12.92 & 22.44
             & 54.97 & 54.60 &  7.12 & 17.22 & 27.65 \\
    Smith    & 25.20 & 30.39 & 25.33 & 53.66 & 69.56
             & 20.87 & 27.59 & 42.54 & 70.62 & 79.22 \\
    \textbf{SMSfP}
             & \textbf{16.99*} & \textbf{23.00*}
             & \textbf{47.56*} & \textbf{80.59*} & \textbf{88.08*}
             & \textbf{13.69*} & \textbf{19.45*}
             & \textbf{59.83*} & \textbf{85.46*} & \textbf{90.58*} \\
    \bottomrule
  \end{tabular}
\end{table}

\begin{figure}[!t]\rmfamily
  \centering
  \includegraphics[width=\textwidth]{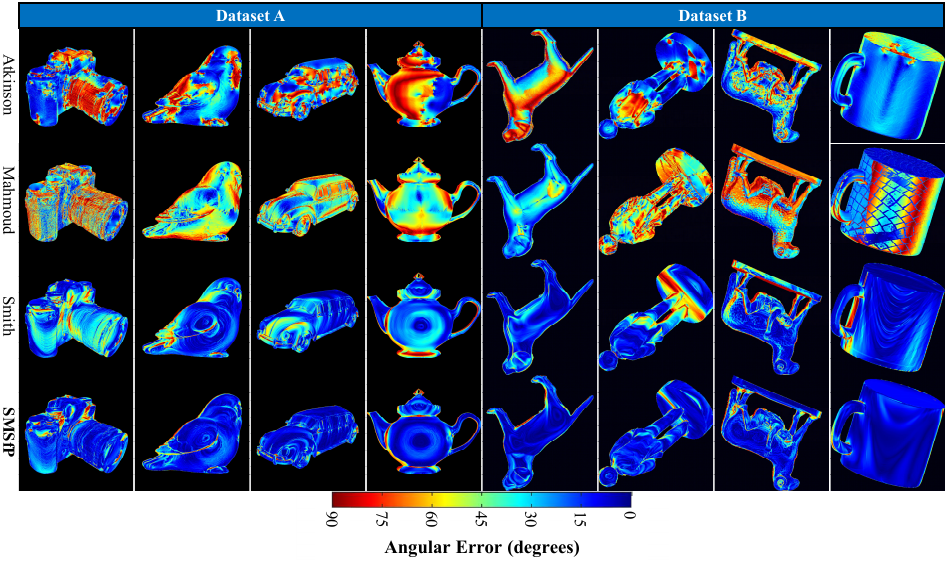}
  \caption{Visual comparison of error maps from different methods on
    dataset A and dataset B.
    From top to bottom, the rows show the results of the methods of
    Atkinson~\cite{ref10}, Mahmoud~\cite{ref12}, Smith~\cite{ref14},
    and the proposed SMSfP method, respectively.
    The color bar on the bottom shows the amount of angular error in
    degrees, where blue means lower error and red means higher error.}
  \label{fig6}
\end{figure}

Figure~\ref{fig6} presents the error analysis comparison.
From top to bottom, each row shows the angular error distribution maps
on dataset A and dataset B of different methods, with colours from blue
to red representing the $0^\circ$--$90^\circ$ error range.
Error analysis reveals: the Atkinson's and Mahmoud's methods exhibit
large red-orange regions indicating severe angular deviations; the
Smith's method shows improvement but still contains considerable error
areas; our method's error maps show significantly reduced angular errors,
indicating merely minor deviations in a few boundary regions.
This comparison intuitively validates the advantages of our method.

\subsection{Ablation Study on Synthetic Datasets}

\begin{figure}[h]\rmfamily
  \centering
  \includegraphics[width=0.75\columnwidth]{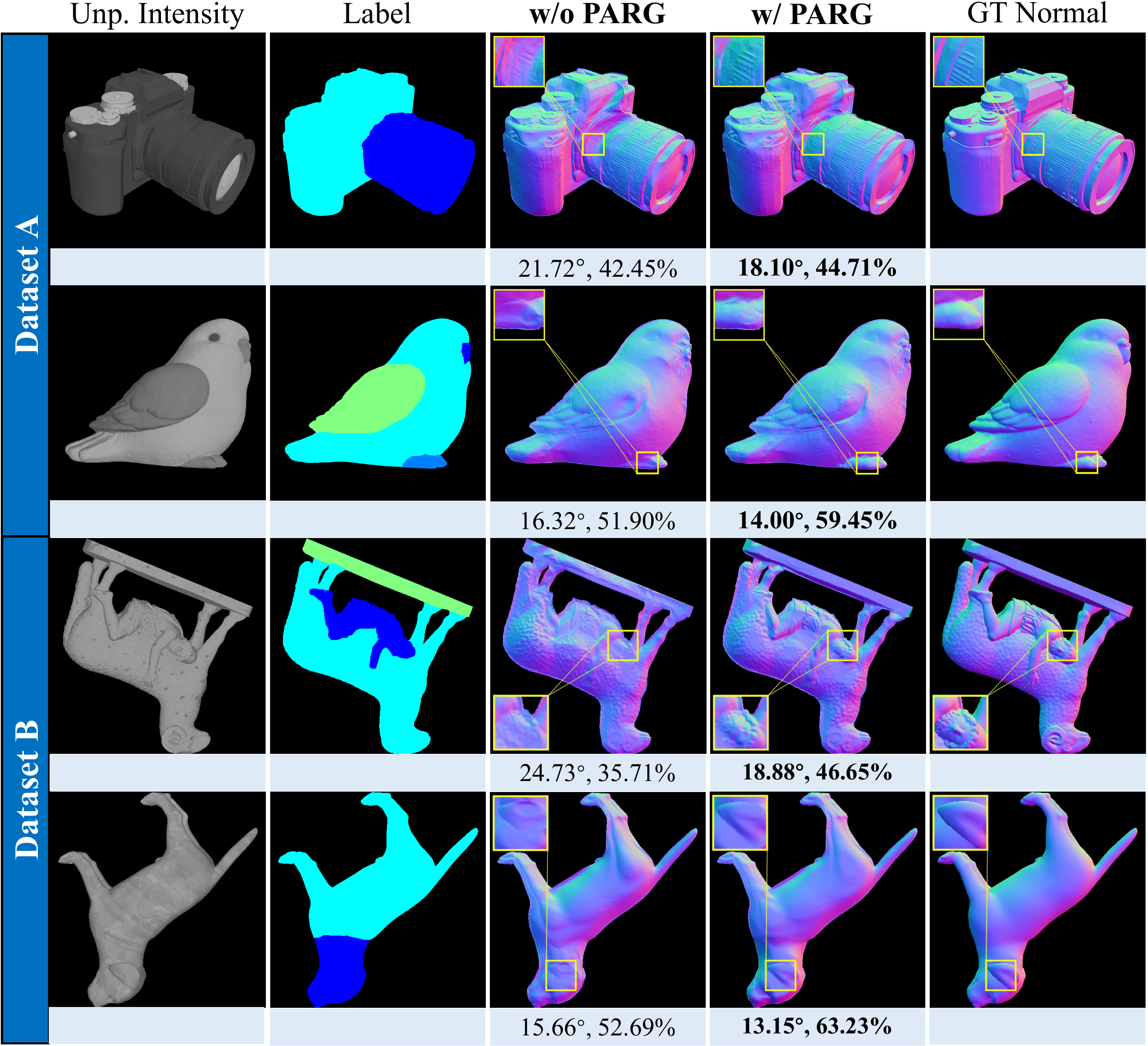}
  \caption{Ablation study comparing the reconstruction results with and without the segmentation module on partial dataset A and dataset B. From left to right: unpolarized intensity images, segmentation labels, reconstruction results without and with PARG segmentation method, and the GT normal maps.} The numbers below each result indicate the MAE and pixel accuracy ($<11.25^\circ$). Zoomed insets highlight the reconstructed local details.
  \label{fig7}
\end{figure}

Figure~\ref{fig7} demonstrates the effectiveness of the PARG segmentation method through an ablation study, where ``w/ PARG'' and ``w/o PARG'' denote the model with and without the PARG segmentation module, respectively. From top to bottom, the rows display the results for the camera, bird, sheep and dog, respectively drawn from dataset A and dataset B. The PARG segmentation method achieves consistent improvements across all test objects, with MAE reductions of
$1.56^\circ$--$5.85^\circ$ and pixel accuracy gains of
3.49\%--10.94\% at the angle threshold of $11.25^\circ$.
Visual comparison reveals that the PARG segmentation method produces notably smoother surface reconstructions with improved geometric consistency. That is because it not only effectively handles the challenging regions with complex convexity, but also preserves fine textural details, thus validating the MFCP constraint for complex surfaces.

Table~\ref{tab2} quantifies the PARG's contribution through the ablation analysis, where ``*'' indicates the best performance under a certain evaluation metric and the performance of SMSfP is indicated in bold. As shown in Table~\ref{tab2}, the results w/o PARG  show notably higher angular errors and lower pixel accuracy across all objects. In contrast, the results w/ PARG achieves consistent improvements across all test objects. On average, the MAE is improved by $3.23^\circ$ (from $20.20^\circ$ to $16.97^\circ$) and the RMSE is reduced by $3.64^\circ$ (from $26.97^\circ$ to $23.33^\circ$), indicating a significant improvement in accuracy. The pixel accuracy is improved by 7.65\%, 5.36\% and 4.93\% at the thresholds of $11.25^\circ$, $22.5^\circ$ and $30^\circ$, respectively. These results confirm that the proposed PARG segmentation method can effectively enhance the reconstruction quality across different object geometries and precision requirements with good robustness.


\begin{table}[htbp]\rmfamily
  \centering
  \caption{Accuracy comparisons on the partial dataset A and dataset B
    for ablation experiments.}
  \label{tab2}
  \footnotesize
  \setlength{\tabcolsep}{4pt}
  \begin{tabular}{@{} l ccccc ccccc @{}}
    \toprule
    \multirow{3}{*}{Object}
      & \multicolumn{5}{c}{w/o PARG}
      & \multicolumn{5}{c}{w/ PARG} \\
    \cmidrule(lr){2-6}\cmidrule(lr){7-11}
      & \multicolumn{2}{c}{Angular Error (deg.)}
      & \multicolumn{3}{c}{Pixel Accuracy (\%)}
      & \multicolumn{2}{c}{Angular Error (deg.)}
      & \multicolumn{3}{c}{Pixel Accuracy (\%)} \\
    \cmidrule(lr){2-3}\cmidrule(lr){4-6}
    \cmidrule(lr){7-8}\cmidrule(lr){9-11}
      & MAE & RMSE & $11.25^\circ$ & $22.5^\circ$ & $30^\circ$
      & MAE & RMSE & $11.25^\circ$ & $22.5^\circ$ & $30^\circ$ \\
    \midrule
    Camera
      & 23.15 & 29.44 & 33.30 & 68.85 & 80.04
      & \textbf{20.19*} & \textbf{26.26*} & \textbf{36.79*} & \textbf{72.27*} & \textbf{84.13*} \\
    Bird
      & 17.23 & 22.78 & 43.64 & 81.30 & 88.30
      & \textbf{15.67*} & \textbf{20.86*} & \textbf{49.29*} & \textbf{84.31*} & \textbf{91.19*} \\
    Sheep
      & 24.73 & 33.33 & 35.71 & 64.70 & 73.32
      & \textbf{18.88*} & \textbf{26.57*} & \textbf{46.65*} & \textbf{75.28*} & \textbf{82.83*} \\
    Dog
      & 15.66 & 22.33 & 52.69 & 80.78 & 87.84
      & \textbf{13.15*} & \textbf{19.63*} & \textbf{63.23*} & \textbf{85.24*} & \textbf{91.06*} \\
    \midrule
    Average
      & 20.20 & 26.97 & 41.34 & 73.91 & 82.38
      & \textbf{16.97*} & \textbf{23.33*} & \textbf{48.99*} & \textbf{79.27*} & \textbf{87.31*} \\
    \bottomrule
  \end{tabular}
\end{table}

\subsection{Real-World Data Acquisition and Validation}

\begin{figure}[htbp]\rmfamily
  \centering
  \includegraphics[width=0.65\columnwidth]{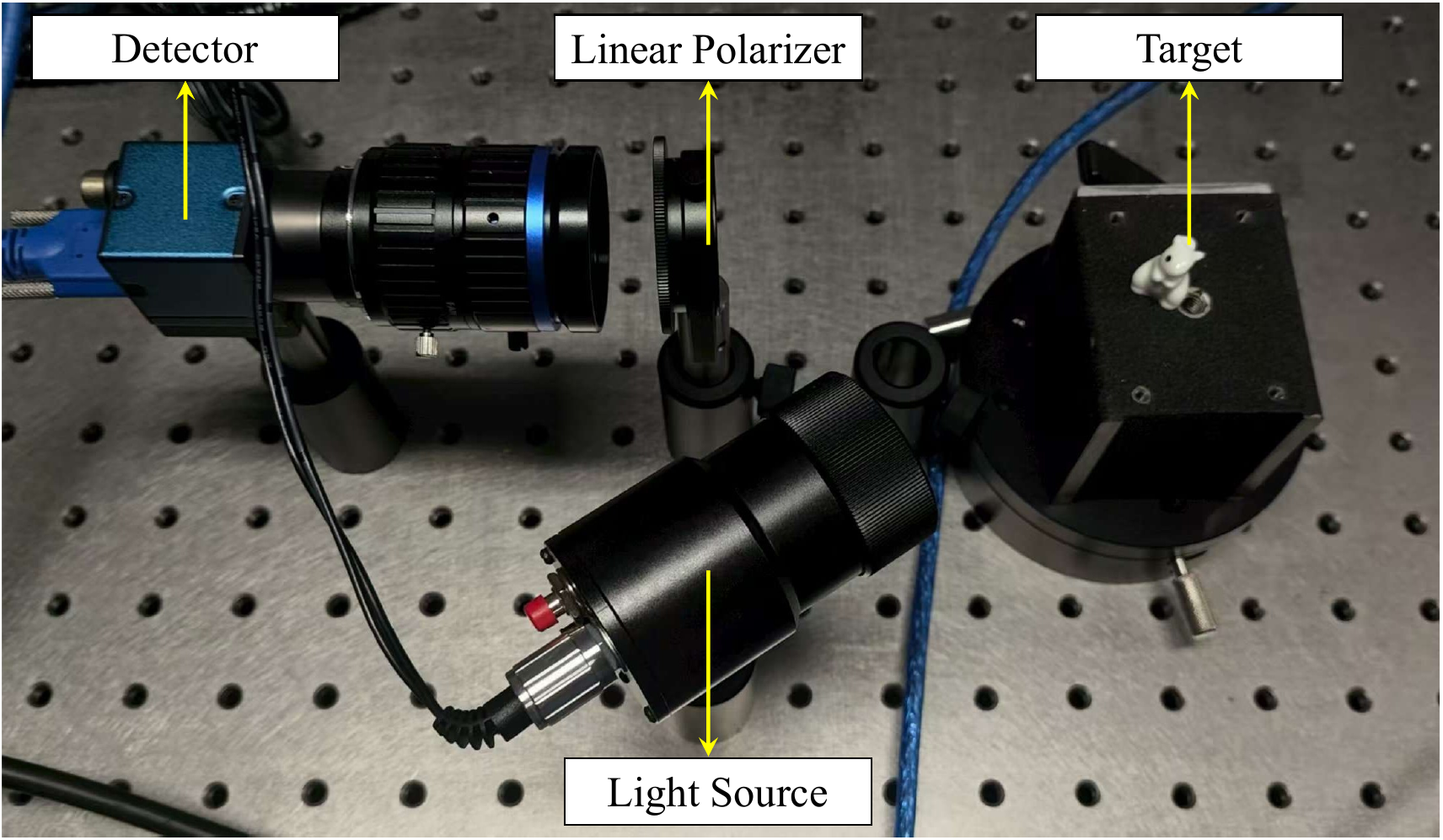}
  \caption{The polarized imaging testbed consisting of the light source,
    target, linear polarizer and detector.}
  \label{fig8}
\end{figure}

To validate the proposed SMSfP method on real-world scenes, we construct
a polarized imaging testbed, as shown in Fig.~\ref{fig8}.
The real-world dataset consists of four miniature models captured at four
polarization angles with $1920\times1200$ resolution.
The system captures images at four distinct polarization angles
($0^\circ$, $45^\circ$, $90^\circ$, and $135^\circ$) by manually
rotating a linear polarizer.
The testbed is composed of four main components: a light source (Daheng
Optics GCI-060411), a detector (Daheng Imaging MER2-231-41U3C), a
linear polarizer (Daheng Optics GCL-050003), and a target object.
All components are aligned along the optical axis and mounted on a
stable optical breadboard.
The system is calibrated with proper focusing and white balance to
ensure image sharpness, colour accuracy, and system stability for
consistent measurements.
The parameter settings used in the following real-world experiments are
consistent with those described in Section~5.1.

\begin{figure}[htbp]\rmfamily
  \centering
  \includegraphics[width=0.8\columnwidth]{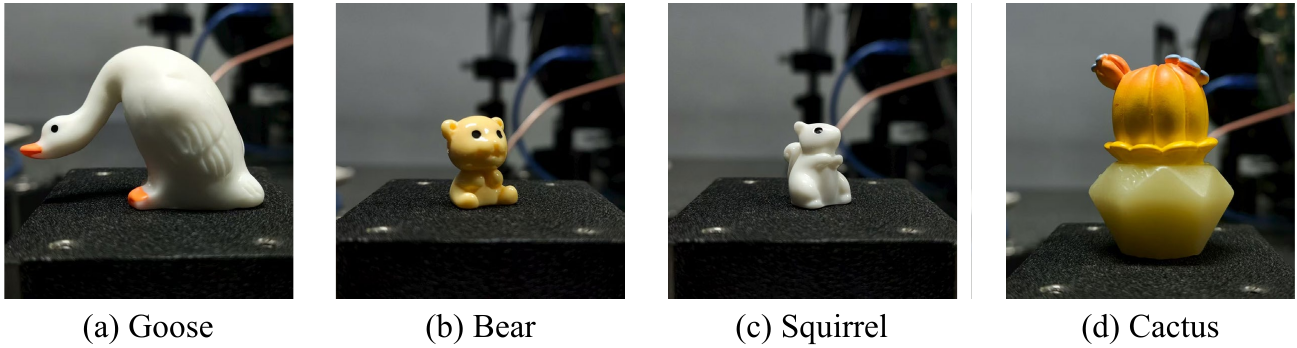}
  \caption{Figures of the four test objects used in the real-world
    experiments: (a)~goose; (b)~bear; (c)~squirrel; (d)~cactus.}
  \label{fig9}
\end{figure}

\begin{figure}[h]\rmfamily
  \centering
  \includegraphics[width=\columnwidth]{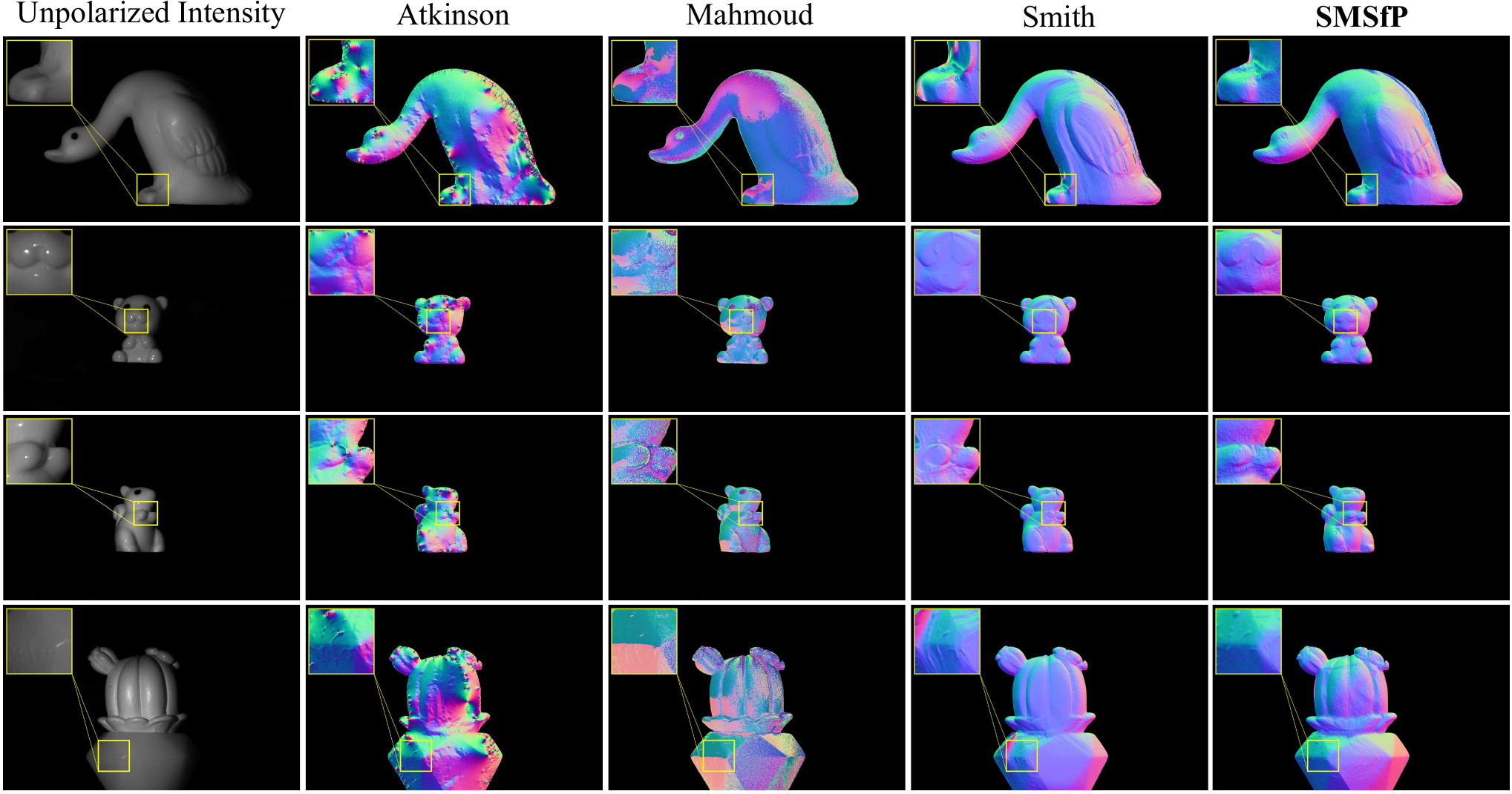}
  \caption{Qualitative comparison on the real-world dataset.
    From top to bottom, the rows show the reconstruction results for the
    four miniature models.
    The left column shows the input unpolarized intensity images for the four miniature models. Subsequent columns display the reconstructed surface normal maps calculated by the methods of Atkinson~\cite{ref10}, Mahmoud~\cite{ref12},
    Smith~\cite{ref14}, and the proposed SMSfP method.
    Zoomed insets are used to highlight the performance on the fine
    geometric details.
    Our method demonstrates superior robustness in preserving smooth
    surface continuity and recovering intricate details compared to the
    baseline approaches.}
  \label{fig10}
\end{figure}

Figure~\ref{fig9} shows the figures of the four test objects (from left
to right: goose, bear, squirrel and cactus) used in the real-world
experiments.
Figure~\ref{fig10} presents the real-world validation results across
four test objects.
Our method consistently outperforms baselines, producing coherent
surface reconstructions with preserved rich details.
While the baseline methods exhibit artefacts and discontinuities,
particularly in the regions with significant variations of surface
curvature, our algorithm maintains smooth surface continuity and
comprehensive coverage.
The zoomed insets highlight these improvements, which demonstrate the
enhanced robustness to the real-world imaging conditions and superior
reconstruction fidelity compared to the traditional methods.

\FloatBarrier

\section{Discussion}

Despite the reconstruction improvements demonstrated across diverse datasets, there are certain considerations regarding the applicability of the proposed SMSfP, where the introduction of PARG segmentation may be counterproductive. The artificially imposed region boundaries degrade the surface continuity and introduce local artefacts, suggesting that selectively applying the segmentation method would better accommodate objects with varying surface topographies.

Additionally, the proposed method assumes purely diffuse reflection, which may not hold in practice. When specular and diffuse reflections coexist on the same surface, the diffuse reflection assumption underlying the polarimetric model no longer holds, which potentially reduces the reconstruction accuracy in specular-dominant regions. Addressing the mixed specular-diffuse reflection scenario represents an important direction for future work.

\section{Conclusion}

This paper developed a novel segmentation-reconstruction method to
overcome the azimuth angle ambiguity in the existing monocular SfP
technology.
The proposed PARG segmentation method transforms the complex global
reconstruction problem into independent locally convex sub-region
reconstructions.
Meanwhile, the MFCP constraint is proposed to preserve the textural
details of reconstructed objects.
Experimental results demonstrated substantial accuracy improvement over
the existing monocular SfP methods across diverse datasets.
By solving the ambiguity problem, the proposed monocular passive system
opens a new window for the low-cost hardware in practical 3D imaging
applications.
Our future research will focus on developing adaptive segmentation strategies for surfaces with varying topographies and addressing the mixed specular-diffuse reflection scenario.


\section{CRediT authorship contribution statement}

\textbf{Jinyu Zhang}: Writing – original draft, Methodology, Validation,
Conceptualization, Investigation. \textbf{Xu Ma}: Writing – review \&
editing, Methodology, Conceptualization, Supervision, Resources.
\textbf{Weili Chen}: Writing – review \& editing, Supervision, Project
administration.

\section{Declaration of competing interest}

The authors declare that they have no known competing financial 
interests or personal relationships that could have appeared to influence 
the work reported in this paper.

\section{Acknowledgment}

Jinyu Zhang and Xu Ma are supported by the National Natural Science
Foundation of China (NSFC) (U2241275).

\section{Data availability}

Data will be made available on request.

\bibliographystyle{cas-model2-names}

\begin{thebibliography}{43}

\bibitem{ref1}
X. Han, T. Li, C. Zheng, Image-based 3D object reconstruction: State-of-the-art and trends in the deep learning era, \emph{IEEE Trans. Pattern Anal. Mach. Intell.} 43 (2019) 1578--1604. \href{https://doi.org/10.1109/tpami.2019.2954885}{\nolinkurl{https://doi.org/10.1109/tpami.2019.2954885}}.

\bibitem{ref2}
M. Contreras, A. Jain, N.P. Bhatt, A. Banerjee, E. Hashemi, A survey on 3D object detection in real time for autonomous driving, \emph{Front. Robot. AI} 11 (2024) 1212070. \href{https://doi.org/10.3389/frobt.2024.1212070}{\nolinkurl{https://doi.org/10.3389/frobt.2024.1212070}}.

\bibitem{ref3}
M. Sarmah, A. Neelima, H.R. Singh, Survey of methods and principles in three-dimensional reconstruction from two-dimensional medical images, \emph{Vis. Comput. Ind. Biomed. Art} 6 (2023) 15. \href{https://doi.org/10.1186/s42492-023-00142-7}{\nolinkurl{https://doi.org/10.1186/s42492-023-00142-7}}.

\bibitem{ref4}
M. Bitzidou, D. Chrysostomou, A. Gasteratos, Multi-camera 3D object reconstruction for industrial automation, in: Proc. 19th Adv. Prod. Manage. Syst. Conf. (APMS 2012), Springer, Rhodes, Greece, 2012, pp. 526--533. \href{https://doi.org/10.1007/978-3-642-40352-1_66}{\nolinkurl{https://doi.org/10.1007/978-3-642-40352-1_66}}.

\bibitem{ref5}
W. Lu, Y. Zhang, X. Chen, M. Zhang, A comprehensive review of vision-based 3D reconstruction methods, \emph{Sensors} 24 (2024) 2314. \href{https://doi.org/10.3390/s24072314}{\nolinkurl{https://doi.org/10.3390/s24072314}}.

\bibitem{ref6}
J. Forest, J. Salvi, E. Cabruja, C. Pous, Structured light and stereo vision for underwater 3D reconstruction, in: OCEANS 2004 MTS/IEEE TECHNO-OCEAN, IEEE, Kobe, Japan, 2004, pp. 1396--1401. \href{https://doi.org/10.1109/oceans-genova.2015.7271433}{\nolinkurl{https://doi.org/10.1109/oceans-genova.2015.7271433}}.

\bibitem{ref7}
X. Li, Z. Liu, Y. Cai, C. Pan, J. Song, J. Wang, X. Shao, Polarization 3D imaging technology: a review, \emph{Front. Phys.} 11 (2023) 1198457. \href{https://doi.org/10.3389/fphy.2023.1198457}{\nolinkurl{https://doi.org/10.3389/fphy.2023.1198457}}.

\bibitem{ref8}
R. Adams, L. Bischof, Seeded region growing, \emph{IEEE Trans. Pattern Anal. Mach. Intell.} 16 (1994) 641--647. \href{https://doi.org/10.1109/34.295913}{\nolinkurl{https://doi.org/10.1109/34.295913}}.

\bibitem{ref9}
L.B. Wolff, Surface orientation from polarization images, in: Proc. SPIE 0850, Optics, Illumination, and Image Sensing for Machine Vision II, SPIE, Cambridge, MA, USA, 1988, pp. 110--121. \href{https://doi.org/10.1117/12.942866}{\nolinkurl{https://doi.org/10.1117/12.942866}}.

\bibitem{ref10}
G.A. Atkinson, E.R. Hancock, Recovery of surface orientation from diffuse polarization, \emph{IEEE Trans. Image Process.} 15 (2006) 1653--1664. \href{https://doi.org/10.1109/tip.2006.871114}{\nolinkurl{https://doi.org/10.1109/tip.2006.871114}}.

\bibitem{ref11}
D. Miyazaki, M. Kagesawa, K. Ikeuchi, Determining shapes of transparent objects from two polarization images, in: Proc. IAPR Workshop on Machine Vision Applications, Nara, Japan, 2002, pp. 26--31.

\bibitem{ref12}
A.H. Mahmoud, M.T. El-Melegy, A.A. Farag, Direct method for shape recovery from polarization and shading, in: Proc. 19th IEEE Int. Conf. Image Process. (ICIP), IEEE, Orlando, FL, USA, 2012, pp. 1769--1772. \href{https://doi.org/10.1109/icip.2012.6467223}{\nolinkurl{https://doi.org/10.1109/icip.2012.6467223}}.

\bibitem{ref13}
W.A.P. Smith, R. Ramamoorthi, S. Tozza, Linear depth estimation from an uncalibrated, monocular polarisation image, in: Computer Vision -- ECCV 2016, Springer, Amsterdam, The Netherlands, 2016, pp. 109--125. \href{https://doi.org/10.1007/978-3-319-46484-8_7}{\nolinkurl{https://doi.org/10.1007/978-3-319-46484-8_7}}.

\bibitem{ref14}
W.A.P. Smith, R. Ramamoorthi, S. Tozza, Height-from-polarisation with unknown lighting or albedo, \emph{IEEE Trans. Pattern Anal. Mach. Intell.} 41 (2019) 2875--2888. \href{https://doi.org/10.1109/tpami.2018.2868065}{\nolinkurl{https://doi.org/10.1109/tpami.2018.2868065}}.

\bibitem{ref15}
T.T. Ngo, H. Nagahara, R. Taniguchi, Shape and light directions from shading and polarization, in: Proc. IEEE Conf. Comput. Vis. Pattern Recognit. (CVPR), IEEE, Boston, MA, USA, 2015, pp. 2310--2318. \href{https://doi.org/10.1109/cvpr.2015.7298844}{\nolinkurl{https://doi.org/10.1109/cvpr.2015.7298844}}.

\bibitem{ref16}
G.A. Atkinson, E.R. Hancock, Surface reconstruction using polarization and photometric stereo, in: Computer Analysis of Images and Patterns (CAIP 2007), Springer, Vienna, Austria, 2007, pp. 1--8. \href{https://doi.org/10.1007/978-3-540-74272-2_58}{\nolinkurl{https://doi.org/10.1007/978-3-540-74272-2_58}}.

\bibitem{ref17}
G.A. Atkinson, E.R. Hancock, Surface shape and reflectance analysis using polarisation, \emph{Comput. Vis. Image Underst.} 142 (2016) 58--69.

\bibitem{ref18}
D. Miyazaki, M. Kagesawa, K. Ikeuchi, Shape from polarization: a method for solving zenithal angle ambiguity, in: Proc. 9th IEEE Int. Conf. Comput. Vis. (ICCV), IEEE, Nice, France, 2003, pp. 1501--1508. \href{https://doi.org/10.1364/ol.37.004218}{\nolinkurl{https://doi.org/10.1364/ol.37.004218}}.

\bibitem{ref19}
S. Rahmann, N. Canterakis, Active lighting applied to three-dimensional reconstruction of specular metallic surfaces by polarization imaging, in: Proc. IEEE Comput. Soc. Conf. Comput. Vis. Pattern Recognit. (CVPR), IEEE, Kauai, HI, USA, 2001, pp. I-149--I-155. \href{https://doi.org/10.1364/ao.45.004062}{\nolinkurl{https://doi.org/10.1364/ao.45.004062}}.

\bibitem{ref20}
C.P. Huynh, A. Robles-Kelly, E.R. Hancock, Uncalibrated, two source photo-polarimetric stereo, in: Computer Vision -- ECCV 2010, Springer, Heraklion, Crete, Greece, 2010, pp. 111--125. \href{https://doi.org/10.1109/tpami.2021.3078101}{\nolinkurl{https://doi.org/10.1109/tpami.2021.3078101}}.

\bibitem{ref21}
Y. Ba, A. Gilbert, F. Wang, J. Yang, R. Chen, Y. Wang, L. Yan, B. Shi, A. Kadambi, Polarized 3D: High-quality depth sensing with polarization cues, in: Computer Vision -- ECCV 2020, Springer, Glasgow, UK, 2020, pp. 558--575. \href{https://doi.org/10.1109/iccv.2015.385}{\nolinkurl{https://doi.org/10.1109/iccv.2015.385}}.

\bibitem{ref22}
Z. Cui, J. Gu, B. Shi, P. Tan, J. Kautz, Polarimetric multi-view stereo, in: Proc. IEEE Conf. Comput. Vis. Pattern Recognit. (CVPR), IEEE, Honolulu, HI, USA, 2017, pp. 1558--1567. \href{https://doi.org/10.1109/CVPR.2017.47}{\nolinkurl{https://doi.org/10.1109/CVPR.2017.47}}.

\bibitem{ref23}
D. Zhu, W.A.P. Smith, Depth from a polarisation + RGB stereo pair, in: Proc. IEEE/CVF Conf. Comput. Vis. Pattern Recognit. (CVPR), IEEE, Long Beach, CA, USA, 2019, pp. 7586--7595. \href{https://doi.org/10.1109/CVPR.2019.00777}{\nolinkurl{https://doi.org/10.1109/CVPR.2019.00777}}.

\bibitem{ref24}
Y. Ba, A. Gilbert, F. Wang, J. Yang, R. Chen, Y. Wang, L. Yan, B. Shi, A. Kadambi, Deep shape from polarization, in: Computer Vision -- ECCV 2020, Springer, Glasgow, UK, 2020, pp. 558--575. \href{https://doi.org/10.1007/978-3-030-58586-0_33}{\nolinkurl{https://doi.org/10.1007/978-3-030-58586-0_33}}.

\bibitem{ref25}
C. Lei, C. Qi, J. Xie, N. Fan, V. Koltun, Q. Chen, Shape from polarization for complex scenes in the wild, in: Proc. IEEE/CVF Conf. Comput. Vis. Pattern Recognit. (CVPR), IEEE, New Orleans, LA, USA, 2022, pp. 12632--12641. \href{https://doi.org/10.1109/cvpr52688.2022.01230}{\nolinkurl{https://doi.org/10.1109/cvpr52688.2022.01230}}.

\bibitem{ref26}
X. Tian, R. Liu, Z. Wang, J. Ma, Learning accurate 3D shape based on stereo polarimetric imaging, \emph{Inf. Fusion} 77 (2022) 19--28. \href{https://doi.org/10.1109/cvpr52729.2023.01658}{\nolinkurl{https://doi.org/10.1109/cvpr52729.2023.01658}}.

\bibitem{ref27}
Y. Cui, P. Sarkar, A. Kadambi, R. Ramamoorthi, Shape from polarization with distant lighting estimation, in: Proc. IEEE/CVF Conf. Comput. Vis. Pattern Recognit. (CVPR), IEEE, Seattle, WA, USA, 2020, pp. 3026--3035. \href{https://doi.org/10.1109/tpami.2023.3298376}{\nolinkurl{https://doi.org/10.1109/tpami.2023.3298376}}.

\bibitem{ref28}
K. Yang, P. Han, R. Gong, M. Xiang, J. Liu, Z. Fan, T. Xi, F. Liu, B. Wang, X. Shao, High-quality 3D shape recovery from scattering scenario via deep polarization neural networks, \emph{Opt. Lasers Eng.} 173 (2024) 107935. \href{https://doi.org/10.2139/ssrn.4502684}{\nolinkurl{https://doi.org/10.2139/ssrn.4502684}}.

\bibitem{ref29}
X. Wu, P. Li, X. Zhang, J. Chen, F. Huang, Three dimensional shape reconstruction via polarization imaging and deep learning, \emph{Sensors} 23 (2023) 4592. \href{https://doi.org/10.3390/s23104592}{\nolinkurl{https://doi.org/10.3390/s23104592}}.

\bibitem{ref30}
Y. Lyu et al., SfPUEL: Shape from polarization under unknown environment light, \emph{Adv. Neural Inf. Process. Syst.} 37 (2024) 97184--97202. \href{https://doi.org/10.52202/079017-3082}{\nolinkurl{https://doi.org/10.52202/079017-3082}}.

\bibitem{ref31}
Z. Wan et al., AttentiveSfP: Leveraging DualPool-Former and attention mechanisms for accurate shape from polarization, \emph{Pattern Recognit.} (2025) 112714. \href{https://doi.org/10.1016/j.patcog.2025.112714}{\nolinkurl{https://doi.org/10.1016/j.patcog.2025.112714}}.

\bibitem{ref32}
K. Li et al., SfP-Underwater: Attention-based shape from polarization for underwater scattering environments, \emph{Opt. Laser Technol.} 192 (2025) 113545. \href{https://doi.org/10.1016/j.optlastec.2025.113545}{\nolinkurl{https://doi.org/10.1016/j.optlastec.2025.113545}}.

\bibitem{ref33}
Z. Wan et al., Shape from polarization based on a polarization representation and sparse self-attention, \emph{Opt. Express} 34 (2026) 3183--3196. \href{https://doi.org/10.1364/oe.584588}{\nolinkurl{https://doi.org/10.1364/oe.584588}}.

\bibitem{ref34}
D. Marr, Vision: A Computational Investigation into the Human Representation and Processing of Visual Information, The MIT Press, Cambridge, MA, USA, 1982. \href{https://doi.org/10.7551/mitpress/9780262514620.001.0001}{\nolinkurl{https://doi.org/10.7551/mitpress/9780262514620.001.0001}}.

\bibitem{ref35}
L. Jin, K. Yamaguchi, M. Watanabe, S. Hira, E. Kondoh, B. Gelloz, Polarization characteristics of scattered light from macroscopically rough surfaces, \emph{Opt. Rev.} 22 (2015) 511--520. \href{https://doi.org/10.1007/s10043-015-0117-2}{\nolinkurl{https://doi.org/10.1007/s10043-015-0117-2}}.

\bibitem{ref36}
R.C. Gonzalez, R.E. Woods, Digital Image Processing, second ed., Prentice Hall, Upper Saddle River, NJ, USA, 2002.

\bibitem{ref37}
D. Paglieroni, Distance transforms: Properties and machine vision applications, \emph{Comput. Vis. Graph. Image Process.} 54 (1992) 57--58. \href{https://doi.org/10.1016/1049-9652(92)90034-U}{\nolinkurl{https://doi.org/10.1016/1049-9652(92)90034-U}}.

\bibitem{ref38}
K. He, J. Sun, X. Tang, Guided image filtering, \emph{IEEE Trans. Pattern Anal. Mach. Intell.} 35 (2013) 1397--1409. \href{https://doi.org/10.1109/TPAMI.2012.213}{\nolinkurl{https://doi.org/10.1109/TPAMI.2012.213}}.

\bibitem{ref39}
T.M. Cover, P.E. Hart, Nearest neighbor pattern classification, \emph{IEEE Trans. Inf. Theory} 13 (1967) 21--27. \href{https://doi.org/10.1109/TIT.1967.1053964}{\nolinkurl{https://doi.org/10.1109/TIT.1967.1053964}}.

\bibitem{ref40}
R.A. Horn, Z. Yang, Rank of a Hadamard product, \emph{Linear Algebra Appl.} 591 (2020) 87--98. \href{https://doi.org/10.1016/j.laa.2020.01.005}{\nolinkurl{https://doi.org/10.1016/j.laa.2020.01.005}}.

\bibitem{ref41}
V. Deschaintre, Y. Lin, A. Ghosh, Deep polarization imaging for 3D shape and SVBRDF acquisition, in: Proc. IEEE/CVF Conf. Comput. Vis. Pattern Recognit. (CVPR), IEEE, Nashville, TN, USA, 2021, pp. 15567--15576. \href{https://doi.org/10.1109/cvpr46437.2021.01531}{\nolinkurl{https://doi.org/10.1109/cvpr46437.2021.01531}}.

\bibitem{ref42}
V. Deschaintre, M. Aittala, F. Durand, G. Drettakis, A. Bousseau, Single-image SVBRDF capture with a rendering-aware deep network, \emph{ACM Trans. Graph.} 37 (2018) 128:1--128:15. \href{https://doi.org/10.1145/3197517.3201378}{\nolinkurl{https://doi.org/10.1145/3197517.3201378}}.

\bibitem{ref43}
X. Wang, D. Fouhey, A. Gupta, Designing deep networks for surface normal estimation, in: Proc. IEEE Conf. Comput. Vis. Pattern Recognit. (CVPR), IEEE, Boston, MA, USA, 2015, pp. 539--547. \href{https://doi.org/10.1109/CVPR.2015.7298652}{\nolinkurl{https://doi.org/10.1109/CVPR.2015.7298652}}.

\end{thebibliography}

\end{document}